\documentclass{article} 
\usepackage{iclr2021_conference,times}

\usepackage{amsmath,amsfonts,bm}

\def\eqref#1{equation~\ref{#1}}

\def\1{\bm{1}}

\DeclareMathAlphabet{\mathsfit}{\encodingdefault}{\sfdefault}{m}{sl}
\SetMathAlphabet{\mathsfit}{bold}{\encodingdefault}{\sfdefault}{bx}{n}

\usepackage{hyperref}
\usepackage{url}
\usepackage{amsmath}
\usepackage{graphicx}
\usepackage[english]{babel}
\usepackage{mathtools}
\usepackage{amssymb}
\usepackage{rawfonts}
\usepackage{oldlfont}
\usepackage[ruled,vlined]{algorithm2e}

\usepackage{booktabs,
            makecell, 
            tabularx}

\newcommand{\vecb}{\boldsymbol} 

\usepackage{verbatim}

\usepackage{titlesec}
\usepackage{physics}

\usepackage{titletoc}
\titlecontents{section}[1.5em]{}{\contentslabel{1.5em}}{\hspace*{-1.5em}}{\titlerule*[8pt]{.}\contentspage}
\usepackage{setspace}
\title{$\beta$-VAE Reproducibility: Challenges and Extensions}

\date{December 2021}

\begin{document}

\maketitle

\begin{abstract}
 $\beta$-VAE is a follow-up technique to variational autoencoders that proposes special weighting of the KL divergence term in the VAE loss to obtain disentangled representations. Unsupervised learning is known to be brittle even on toy datasets and a meaningful, mathematically precise definition of disentanglement remains difficult to find. Here we investigate the original $\beta$-VAE paper and add evidence to the results previously obtained indicating its lack of reproducibility. We also further expand the experimentation of the models and include further more complex datasets in the analysis. We also implement an FID scoring metric for the $\beta$-VAE model and conclude a qualitative analysis of the results obtained. We end with a brief discussion on possible future investigations that can be conducted to add more robustness to the claims.
\end{abstract}

\makeatletter{\renewcommand*{\@makefnmark}{}\footnotetext{\textsuperscript{1}University of Oxford; Corresponding author: munib.mesinovic@jesus.ox.ac.uk}}


\section{Introduction}
Variational autoencoders \citep{DBLP:journals/corr/KingmaW13} are a class of unsupervised representation learning models with a principled probabilistic interpretation that extends normal autoencoders first described by \cite{hinton2006reducing}. $\beta$-VAE is a follow-up technique that proposes special weighting of the KL divergence term in the VAE loss to obtain disentangled representations. However, unsupervised learning is notoriously brittle even on toy datasets and a meaningful, mathematically precise definition of disentanglement remains difficult to find.

It is thus not obvious to what extent $\beta$-VAEs can robustly obtain disentangled representations in different settings. The main contributions of our reproducibility report can be summarised as follows:
\begin{enumerate}
    \item We add to the evidence provided by follow-up work \citep{pmlr-v80-kim18b, locatello2020sober} that the almost perfect performance presented by \cite{higgins2016beta} is very difficult to reproduce.
    \item We demonstrate that $\beta>1$ does not continue to yield the best quantitative disentanglement results for very complex datasets.
    \item We show that high disentanglement metric scores do not imply a qualitative disentanglement.
    \item We quantitatively assess how lower $\beta$ values give better reconstructions of the original images.
\end{enumerate}

\subsection{VAE framework}
VAEs are a special class of deep generative models optimised via variational inference, which allows one to approximate intractable distributions in Bayesian inference by solving an optimisation problem. Assume we have a directed latent variable model
\begin{equation}
    p(\vecb{x},\vecb{z}) = p(\vecb{x}|\vecb{z})p(\vecb{z})
\end{equation}
and we observed a dataset $D = \{x_1, x_2, .., x_n\}$. Many standard techniques such as Expectation-Maximization do not scale to the large-scale deep learning setting because they require computing the $p_{\theta}(\vecb{z}|\vecb{x})$, for which the normalization constant is not available. We can avoid the need for precise normalization constants by using a variational approximation $q_{\phi}(\vecb{z}|\vecb{x})$ and instead optimising the evidence lower bound (ELBO) as in Equation (\ref{eq:elbo}), where $p_{\theta}(\vecb{x})$ is the marginal likelihood or model evidence:
\begin{align}
    \label{eq:elbo}
    \text{ELBO} &=  {\mathbb E}_{q_{\phi} (\vecb{z}| \vecb{x})}\left[\log\frac{p_{\theta}(\vecb{z}, \vecb{x})}{q_{\phi}(\vecb{z} | \vecb{x})}\right] \\
    &= {\mathbb E}_{q_{\phi} (\vecb{z}| \vecb{x})}\left[\log p_\theta (\vecb{x} | \vecb{z})\right] - \text{KL}(q_{\phi}(\vecb{z} |\vecb{x}) || p_{\theta}(\vecb{z}))\\
    &= \log p_{\theta}(\vecb{x}) - \text{KL}(q_{\phi}(\vecb{z} |\vecb{x}) || p_{\theta}(\vecb{z}|\vecb{x})) \\
    &\leq \log p_{\theta}(\vecb{x})
\end{align}

However, obtaining gradients of the ELBO with respect to the variational parameters is difficult, because we cannot safely exchange derivatives and integrals in this case. Instead, VAEs crucially rely on using the reparameterization trick for computing Monte Carlo estimates of the gradient, typically using a single minibatch. The reparameterization trick is crucial, since it produces an estimator with much lower variance than more general-purpose Monte Carlo estimators, such as the score function estimator. However, reparameterization requires working with continuous distributions and makes VAEs difficult to apply in the discrete setting, albeit not impossible \citep{DBLP:conf/iclr/MaddisonMT17, NIPS2017_7a98af17}. Standard VAEs typically use an isotropic Gaussian prior for the KL divergence, which enables computing the KL divergence analytically.

\subsection{$\beta$-VAE improvements}
\cite{higgins2016beta} propose to increase the weighting of the KL divergence in Equation (\ref{eq:elbo}). This should in turn enforce a greater similarity between the posterior $p(\vecb{z}|\vecb{x})$ and the prior $p_{\phi}(\vecb{z})$, which leads to greater disentanglement. \cite{bengio2013representation} define disentangled representations as the property that latent variables are sensitive to one of the ground truth generative factors, but invariant to others. Because the VAE prior is typically chosen to be a Gaussian with diagonal covariance matrix, its dimensions are independent, which can be seen as disentangled. 

While the $\beta$-weighted KL loss can be seen merely as a heuristic addition to the normal autoencoder, follow-up work exploiting the information-theoretic nature of KL divergence has led to further improvements in the algorithm \citep{NEURIPS2018_1ee3dfcd, DBLP:journals/corr/abs-1804-03599}. Regardless, large-scale reproduction studies show the importance of random seeds and hyperparameter settings to be at least comparable to model choice \citep{locatello2020sober}.

\section{Methodology}

\subsection{Datasets} \label{subsec:datasets}
\cite{higgins2016beta} use a number of standard image datasets to evaluate both image generation properties of $\beta$-VAEs and specially designed datasets for evaluating disentanglement of ground truth generative factors. 

\subsubsection{Disentanglement evaluation - 2Dshapes, 3Dshapes, MPI3DToy}
\begin{figure}[htp]

\centering
\includegraphics[width=.3\textwidth]{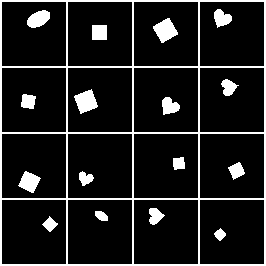}\hfill
\includegraphics[width=.3\textwidth]{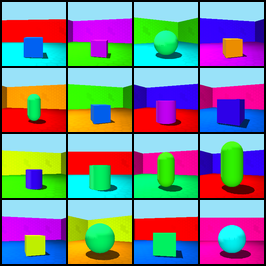}\hfill
\includegraphics[width=.3\textwidth]{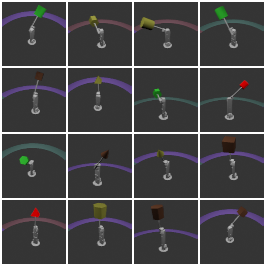}

\caption{Samples from the datasets 2Dshapes, 3Dshapes and MPI3DToy}
\label{fig:dataset_samples}

\end{figure}

To assess the disentanglement quantitatively, we use three synthetic datasets that come with ground truth generative factors, with samples of each shown in Figure \ref{fig:dataset_samples}. 2Dshapes was originally created by \cite{higgins2016beta}, consisting of 737,280 2D shapes that are generated from the Cartesian product of five ground truth independent latent factors. 

We hypothesise that the 2Dshapes dataset is too easy to solve, as evidenced by the fairly high scores of PCA and ICA, and further check the robustness of $\beta$-VAEs on harder, more recent datasets; specifically, RGB datasets 3Dshapes \citep{3Dshapes18} and MPI3DToy \citep{gondal2019transfer}. 3Dshapes consists of 480,000 images generated from six data generative factors. MPI3DToy is the most complex dataset containing 1,036,800 images from seven data generative factors captured from real-world robotics experiments. Due to computational constraints, we only use the toy version of this dataset which contains renders of the real scenes, rather than the scenes themselves.

\subsubsection{Image generation - CelebA, Chairs, CIFAR10, and CIFAR100}
\begin{figure}[htp]

\centering
\includegraphics[width=.3\textwidth]{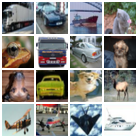}
\includegraphics[width=.3\textwidth]{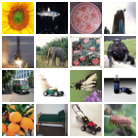}

\caption{Samples from the CIFAR10 and CIFAR100 datasets}
\label{fig:CIFARdataset_samples}

\end{figure}

CelebA and Chairs are the two datasets utilised by \cite{higgins2016beta} for qualitative evaluation. Due to hardware constraints, we instead qualitatively inspect models trained on CIFAR10 and CIFAR100, as they are standard benchmark datasets for evaluating generative models \citep{shmelkov2018good}. These datasets are non-synthetic, and samples are shown in Figure~\ref{fig:CIFARdataset_samples}.  

\subsection{Metrics} \label{subsec:metrics}
To evaluate the disentanglement of a representation, we adopt the approach presented in \cite{higgins2016beta}. This can be used to evaluate disentanglement directly without having to rely on qualitative inspection.

\cite{higgins2016beta} suggest that there exists a trade-off between generated image quality and level of disentanglement. This is at odds with the notion that disentangled representations should lead to superior performance in downstream tasks. To quantitatively investigate the extent to which higher $\beta$ harms generative model quality while enhancing disentanglement, we adopt Fréchet Inception Distance as a state-of-the-art metric for evaluating reconstruction quality of generative models \citep{heusel2017gans}. \cite{higgins2016beta} only evaluate image quality qualitatively by inspection.

\subsubsection{Fréchet Inception Distance}

Fréchet Inception Distance (FID) is a metric used to assess the reconstruction quality of images produced by generative models, usually a generative adversarial network (GAN). Rather than comparing generated and real images on a pixel-by-pixel basis, the FID compares the distribution of the activations of the final layer of a pretrained InceptionV3 model \citep{szegedy2016rethinking}. This layer corresponds to high-level features of objects (such as airplanes) and thus captures the human notion of similarity in images \citep{heusel2017gans}. FID is based on the Wasserstein metric and can be computed as:

\begin{equation}
    \text{FID} = \lVert\vecb{\mu}_r - \vecb{\mu}_g\rVert^2 + \text{Tr} (\vecb{\Sigma}_r + \vecb{\Sigma}_g - 2 (\vecb{\Sigma}_r \vecb{\Sigma}_g)^{1/2})
\end{equation}

where \[X\textsubscript{r} \sim {\mathcal{N}}(\vecb{\mu}\textsubscript{r},\,\vecb{\Sigma}_r)\] and \[X\textsubscript{g} \sim {\mathcal{N}}(\vecb{\mu}\textsubscript{g},\,\vecb{\Sigma}_g)\] are Gaussians fit to the 2048-dimensional activations of the last InceptionV3 pooling layer for real and generated samples, respectively \citep{FrechetI11}. As we used a pretrained model, images need to be scaled to the correct size and have colour channels added if they were black and white.

We use FID to examine the trade-off between disentanglement and reconstruction quality since the best performing models for image generation (i.e. GANs) use FID for evaluation \citep{dai2019diagnosing, lucic2017gans}. A lower FID score corresponds to real and generated samples being more similar. WGAN-GP \citep{NIPS2017_892c3b1c}, a GAN of similar age as the $\beta$-VAE, is noted to achieve an FID of 29.3 on CIFAR10 \citep{heusel2017gans}. Early iterations of VAEs are generally understood to have poor image generation capabilities compared to contemporary GANs. \citep{shmelkov2018good, makhzani2015adversarial}. However, recent VAE-based models \citep{parmar2020dual} achieve a CIFAR10 FID of 17.9, and are competitive with GAN-based approaches.

Limitations when using FID as a metric are indicated by \cite{shmelkov2018good}. FID cannot separate image quality from image diversity. For example, a poor FID score can be due to the reconstructed images either being unrealistic (low image quality) or too similar to each other (low diversity), with no way to analyze the cause.

\subsubsection{Disentanglement Metric}
The disentanglement metric refers to the framework proposed by \cite{higgins2016beta} that is meant to quantify the level of disentanglement in deep generative models by measuring the independence and interpretability of their latent representation. The idea is that for a disentangled representation, images generated from fixing one factor of variation $y$ and randomly sampling all others should result in a relatively lower variance in the latents corresponding to $y$. The lower this variance is, the easier it will be to predict the corresponding data generative factor. Therefore, we can measure the disentanglement by reporting the accuracy of a classifier identifying the corresponding data generative factor given the latent representation.

An assumption on the dataset $X$ is that its elements ${\vecb{x}} \in {\mathbb R}^N$ are generated from a true world simulator 
\begin{equation}
    p(\vecb{x} | \vecb{v}, \vecb{w}) = {\mathbf{Sim}}(\vecb{v}, \vecb{w}),
\end{equation}
where ${\vecb{v}}\in {\mathbb{R}}^K$ with $\log p({\vecb{v}} | {\vecb{x}}) = \sum_k \log p(v_k| {\vecb{x}})$ are conditionally independent factors  and ${\vecb{w}}\in {\mathbb R}^H$ are conditionally dependent factors. In particular, these ground truth factors need to be known for computing the metric. In all of the datasets that we consider in this report for computing the metric, the data is generated by independent factors $\vecb{v}$ alone. The full procedure is outlined in algorithm \ref{algo:DMetric}.

\begin{algorithm}[h]
\SetAlgoLined
\KwResult{Classification accuracy}
 Uniformly sample one of the given data generative factors $y\sim Unif[1, \ldots, K]$\;
 \For{$b \in [1, \ldots, B]$}{
     \For{$l \in [0, \ldots, L]$}{
      Sample a pair ${\vecb{v}}_{1,l}, {\vecb{v}}_{2,l}$ such that they agree on their $y$th value $\vecb{v}_{1,l}^y = \vecb{v}_{2,l}^y$\;
      Simulate images ${\vecb{x}}_{1,l} \sim {\mathbf{Sim}}({\vecb{v}}_{1,l})$, ${\vecb{x}}_{2,l} \sim {\mathbf{Sim}}({\vecb{v}}_{2,l})$\;
      Infer the expectation values of the latent Gaussians ${\vecb{z}}_{1,l} = {\vecb{\mu}}({\vecb{x}}_{1,l})$, ${\vecb{z}}_{2,l} = {\vecb{\mu}}({\vecb{x}}_{2,l})$\;
      Compute elementwise differences and absolute values ${\vecb{z}}_\text{diff}^ l = |{\vecb{z}}_{1,l} - {\vecb{z}}_{2,l}|$ \;
     }
     ${\vecb{z}}_\text{diff}^b = \frac{1}{L} \sum_l  {\vecb{z}}_\text{diff}^ l$
 }
 Train a classifier using the inputs ${\vecb{z}}_\text{diff}^b$ to predict $y$\;
 Report classification accuracy on a test set.
 \label{algo:DMetric}
 \caption{The Disentanglement Metric \citep{higgins2016beta}}
\end{algorithm}
Note that rather than using the expectation values of the latent Gaussian for one simulated image, a pair of images ${\vecb{x}}_{1,l}, {\vecb{x}}_{2,l}$ is sampled, and the absolute difference of their latent representations is computed to reduce the variance of the classifier's inputs and to lower the conditional dependence on the input images ${\vecb{x}}_{1,l}, {\vecb{x}}_{2,l}$. The classifier is taken to be linear to ensure the interpretability of the inferred latents and not learn any nonlinear disentanglement itself.

One drawback of this metric is its dependence on hyperparameters, such as the choice of the classifier, the optimiser, and most significantly the sample size $L$. This is already noted by \cite{pmlr-v80-kim18b}. We investigate this further by reporting the accuracy scores of a linear and nonlinear MLP, a logistic regression classifier and a random forest classifier from the \emph{SKlearn} library. Furthermore, we compare the scores for different values of the sample size $L$ in generating training data for the metric.

In addition, \cite{pmlr-v80-kim18b} show that there is a mode in which the classifier reports 100\% accuracy while only $K-1$ factors are disentangled. Issues like this could account for the discrepancy between quantitative scores and qualitative observations as described in Section~\ref{subsubsection:quantvsqual}.

\subsection{Models}
\cite{higgins2016beta} originally use a VAE where both the encoder and decoder are MLPs for the 2Dshapes dataset and a Convolutional VAE for their remaining experiments. Follow-up work \citep{DBLP:journals/corr/abs-1804-03599, locatello2020sober, pmlr-v80-kim18b} instead uses a Convolutional VAE across all experiments, including 2Dshapes.

We report results for both the MLP and Convolutional VAEs on 2Dshapes and only use the Convolutional VAE for other experiments as we found it to give better results. As optimisers, we use Adagrad (lr = 1e-2) for the MLP and Adam (lr = 5e-4) for the Convolutional VAE as done in the original papers. We also use PCA and ICA as non-deep baselines, following \cite{higgins2016beta}. All models are trained with multiple seeds. Full details on the model, hyperparameters, and training protocols are available in Appendix \ref{subsec:arch_details}.

\section{Results}
In this section, we present the results using the datasets described in Section \ref{subsec:datasets} and the metrics from Section \ref{subsec:metrics} to evaluate $\beta$-VAEs in terms of their ability to learn a disentangled representation and to reconstruct images.

\subsection{Disentanglement}
First, we present the scores of the disentanglement metric on the dataset used by \cite{higgins2016beta} and compare them to baseline methods PCA and ICA. Next, we investigate the behaviour of $\beta$-VAE on the more complex three-dimensional datasets 3Dshapes and MPI3DToy. Finally, we draw a connection between quantitative and qualitative evaluation of disentanglement to see if these two notions coincide. 

\subsubsection{Disentanglement on 2Dshapes}
Table \ref{tab:dis_metric_2Dshapes} presents the disentanglement scores achieved on the 2Dshapes dataset. It is important to note that among the total five ground truth factors, \cite{higgins2016beta} disregard `shape' when sampling data generative factors for the disentanglement metric. This is presumably because $\beta$-VAE struggles to learn a disentangled representation for this data generative factor, as evidenced by the traversals in Figure 7 from \cite{higgins2016beta} (reprinted in Figure \ref{fig:qual_quant_contrast}) all having a very similar shape.

\begin{table}[h]
    \centering
    
    \begin{tabularx}{0.981\textwidth}{l
                                    r
                                    r 
                                    r
                                    r 
                                    r
                                 }
         \toprule
        & \multicolumn{2}{c}{5 factors} & \multicolumn{3}{c}{4 factors}  \\
        \cmidrule(r){2-3} \cmidrule(l){4-6}
         \bf{Model}  &  \multicolumn{1}{c}{\bf{Mean}}   & \bf{Median} &  \multicolumn{1}{c}{\bf{Mean}} & \bf{Median} & \bf{\cite{higgins2016beta}}  \\
            \midrule
        $0.1$-VAE   &       65.13 $\pm$ 4.98\%   &  64.79\%  & {81.89} $\pm$ 2.33\%    & 81.82\%   & (--)\\
                                
        $1$-VAE              & 64.31$\pm$ 6.56\%  &  63.79\% &  {82.39} $\pm$5.79\%  & 81.37\% & 61.58 $\pm$ 0.5\%\\
                                
        $4$-VAE             & 74.53 $\pm$ 15.06\%  &  76.62\%  &  {89.09} $\pm$ 15.82\% &  95.15\% & 99.23 $\pm$ 0.1\% \\
                                
        $10$-VAE            & 76.77 $\pm$ 11.44\%  & 79.67\%   &  {86.86} $\pm$ 15.5\%  &   88.81\% & (--) \\
                                
            \addlinespace
        PCA                 &  69.79 $\pm$ 4.22\% &  71.53\%  &  {88.95} $\pm$ 3.89\%    & 89.91\% & 84.9 $\pm$ 0.4\% \\
                                
            \addlinespace
        ICA                 &  68.99 $\pm$ 1.68\%  & 69.38\%  & {83.94} $\pm$2.16\%    & 84.12\% & 42.03 $\pm$ 10.6\% \\
                                
            \bottomrule
    \end{tabularx}
    \caption{Disentanglement metric scores on 2Dshapes for $\beta$-VAE, $\beta\in \{0.1,1,4,10\}$, PCA and ICA.}
    \label{tab:dis_metric_2Dshapes}
\end{table}
In Table \ref{tab:dis_metric_2Dshapes} we compare the metric scores using all five data generative factors to the scores obtained by disregarding the `shape' data generative factor. Furthermore, we reprint the results presented in \cite{higgins2016beta} for comparison. We see that the scores using all five data generative factors are lower across all models confirming the conjecture mentioned above. While our results show the superiority of $\beta$-VAE over regular VAE for $\beta >1$, we do not obtain a difference as big as that in \cite{higgins2016beta}. A potential reason for this may be that \cite{higgins2016beta} discard the worst performing 50\% of their training runs due to training instabilities. We also observe this instability especially for $\beta>1$, as suggested by the high standard deviations. In the rest of this report, we report median by default as a robust measure of performance even in case some of the training runs diverge. Furthermore, we note that ICA has a much stronger performance in our results. We found that fine-tuning ICA parameters was crucial for improving its scores.


\subsubsection{Beyond 2D datasets}
\begin{figure}[htp]
\centering
\includegraphics[width=0.8\textwidth]{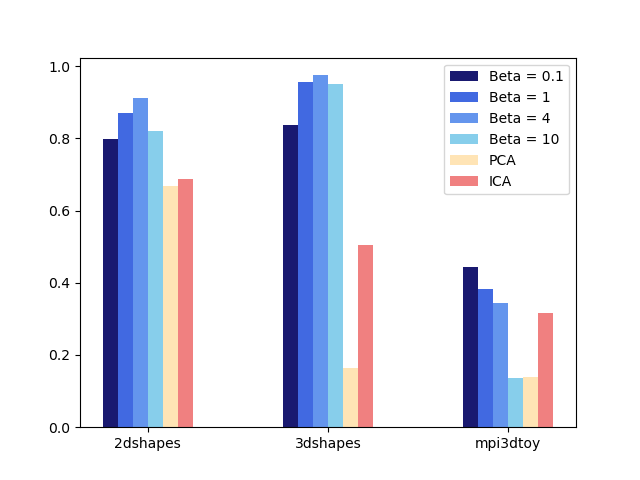}
\caption{Final median disentanglement scores using the Conv. VAE architecture from \cite{burgess2018understanding} for different values of $\beta$, PCA and ICA on the datasets 2Dshapes, 3Dshapes and MPI3DToy.}
\label{fig:DM_Burgess}
\end{figure}

Next, we investigate the disentanglement scores on the more complex datasets 3Dshapes and MPI3DToy. Because images in these datasets have three colour channels, a convolutional architecture for the encoder and decoder is more suitable. Therefore, we trained $\beta$-VAE using the architecture proposed in \cite{burgess2018understanding}, which is also used by follow-up work on disentangling in VAEs, on all of the three datasets with ground truth variation factors. For 2Dshapes in particular, it leads to significantly improved disentanglement scores, which often reach the almost 100$\%$ accuracy reported by \cite{higgins2016beta} when using only four latents on 2Dshapes and around 90$\%$ for all five. These Convolutional VAE experiments are illustrated in Figure \ref{fig:DM_Burgess}, including the five latent 2Dshapes.

$\beta$-VAE manages to reach very high scores on 3Dshapes as well. However, PCA and ICA experience a significant drop in the disentanglement metric. This may be due to the fact that we flattened the 3-channel images to be in a suitable shape for these methods. On the most complex dataset MPI3DToy we see that the best scores are actually reached by $\beta = 0.1$ followed by $\beta =1$.

Compared to MPI3DToy, the 3Dshapes dataset has much higher contrasts and the shapes are more regular, as illustrated in Figure \ref{fig:dataset_samples}. Furthermore, it is generated from six factors of variation compared to the seven of MPI3DToy. We hypothesise that the reason why the lowest $\beta$ performs the best is that since MPI3DToy is the hardest dataset and it has the highest relative magnitude of reconstruction loss to the KL loss. This implicitly creates a different weighting of the two loss terms, suggesting that $\beta$ needs to be scaled according to the difficulty of the dataset.

\subsubsection{Quantitative vs qualitative evaluation of disentanglement}
\label{subsubsection:quantvsqual}
\begin{figure}[htp]
\centering
\includegraphics[width=1\textwidth]{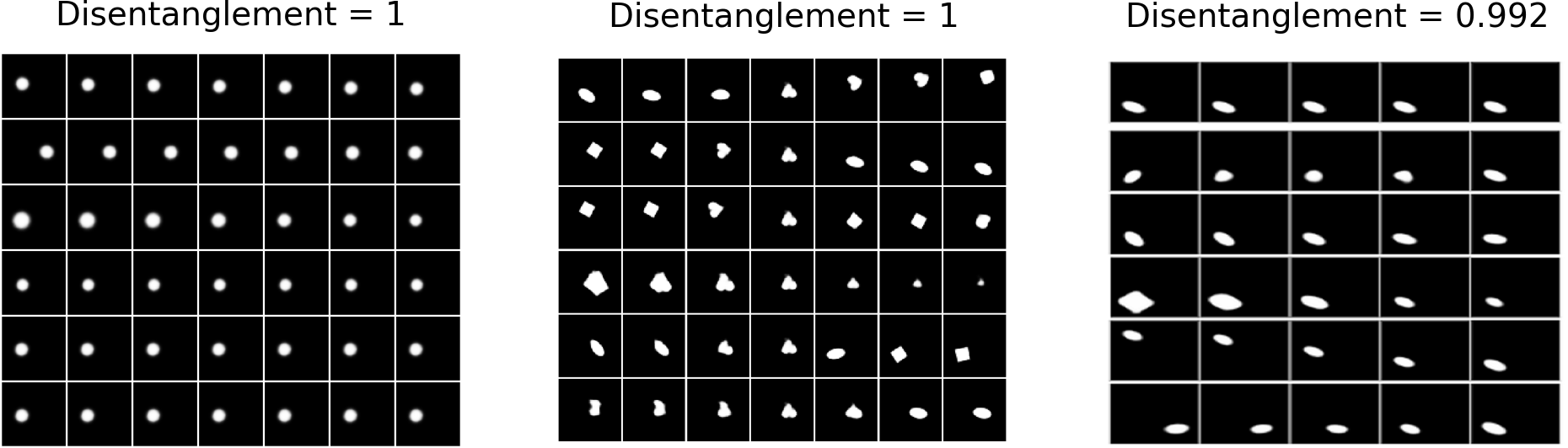}
\caption{Example latent traversals on 2Dshapes from our MLP VAE (left, \cite{higgins2016beta}), our Convolutional VAE (middle, \cite{DBLP:journals/corr/abs-1804-03599}) and the original MLP VAE (right, reprinted from \cite{higgins2016beta}). Latent traversals are done across columns. While the disentanglement scores are very high, qualitative evaluation suggests that the level of disentanglement is quite low.}
\label{fig:qual_quant_contrast}

\end{figure}

The Higgins disentanglement metric tends to assign very high scores that do not correlate well with human judgement of the level of disentanglement. In Figure \ref{fig:qual_quant_contrast}, we first encode the latents of a true data sample from 2Dshapes and then vary the latent dimensions individually. The true generative factors are position X, position Y, scale, rotation and shape. All models achieve almost perfect disentanglement quantitatively, but upon visual inspection, generally only two out of five latents are learnt well and even those might be entangled. 

\subsubsection{Latent space visualization}
\begin{figure}[htp]
\centering
\includegraphics[width=0.7\textwidth]{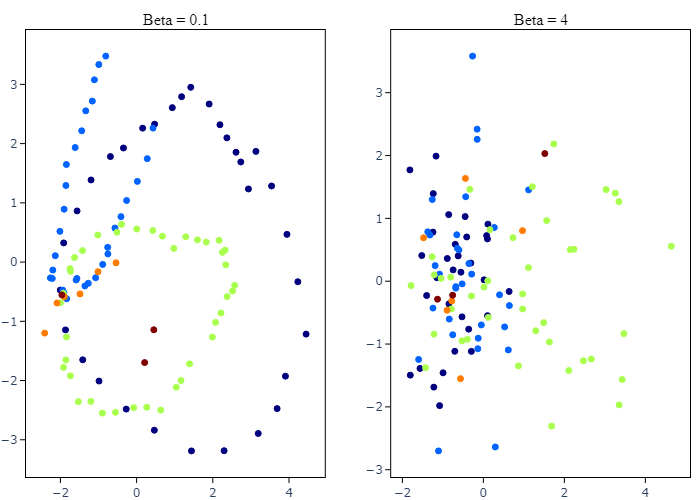}
\caption{Dimensionality reduction of the latent space to the first two components of PCA. Each colour corresponds to variation of one of latent factor. Higher $\beta$ results in denser embeddings of the latent factors.}
\label{fig:gt_latents}
\end{figure}

Figure \ref{fig:gt_latents} shows posterior latents on the 2Dshapes dataset embedded by the first two components of PCA. In order to generate the data, we first seeded the VAE with a picture that had the median values for each of the five ground truth generative factors. Then we conducted a traversal across each dimension in the ground truth latent space (rather than the model posterior latents as in Figure \ref{fig:qual_quant_contrast}).

When the $\beta$ parameter is set low, the embeddings of the ground truth traversal are sparse and appear to have much more regular structure than for higher $\beta$s. This is likely because the KL divergence with a Normal prior penalizes embeddings that are far away from the mean since Normal distributions have thin tails. As a result, the model posterior latents become much more concentrated near the mean and do not extend very far into the latent space.

\cite{spinner_towards_2018} explain how low $\beta$ makes the embeddings more similar to those of a standard autoencoder. This suggests that the posterior latent distribution of standard autoencoders (or VAEs with low $\beta$ for that matter) is strongly concentrated on the training samples. Sampling new images is difficult because decoding a randomly sampled latent code in the autoencoder case is unlikely to generate meaningful images because we are almost certainly not going to hit the part of the latent space that the model learnt to decode during training. However, the KL divergence regularization for $\beta \geq 1$ causes the model to utilise the whole latent space rather than a small part of it. This in turn makes the model work better for generating new data. In Section \ref{subsec:vae_mse}, we formalise the intuition of $\beta$ controlling the posterior latent variance when assuming Gaussian likelihood over the decoded pixels.

\subsubsection{Sensitivity analysis of the disentanglement metric}
\begin{figure}[htp]
\centering
\includegraphics[width=0.8\textwidth]{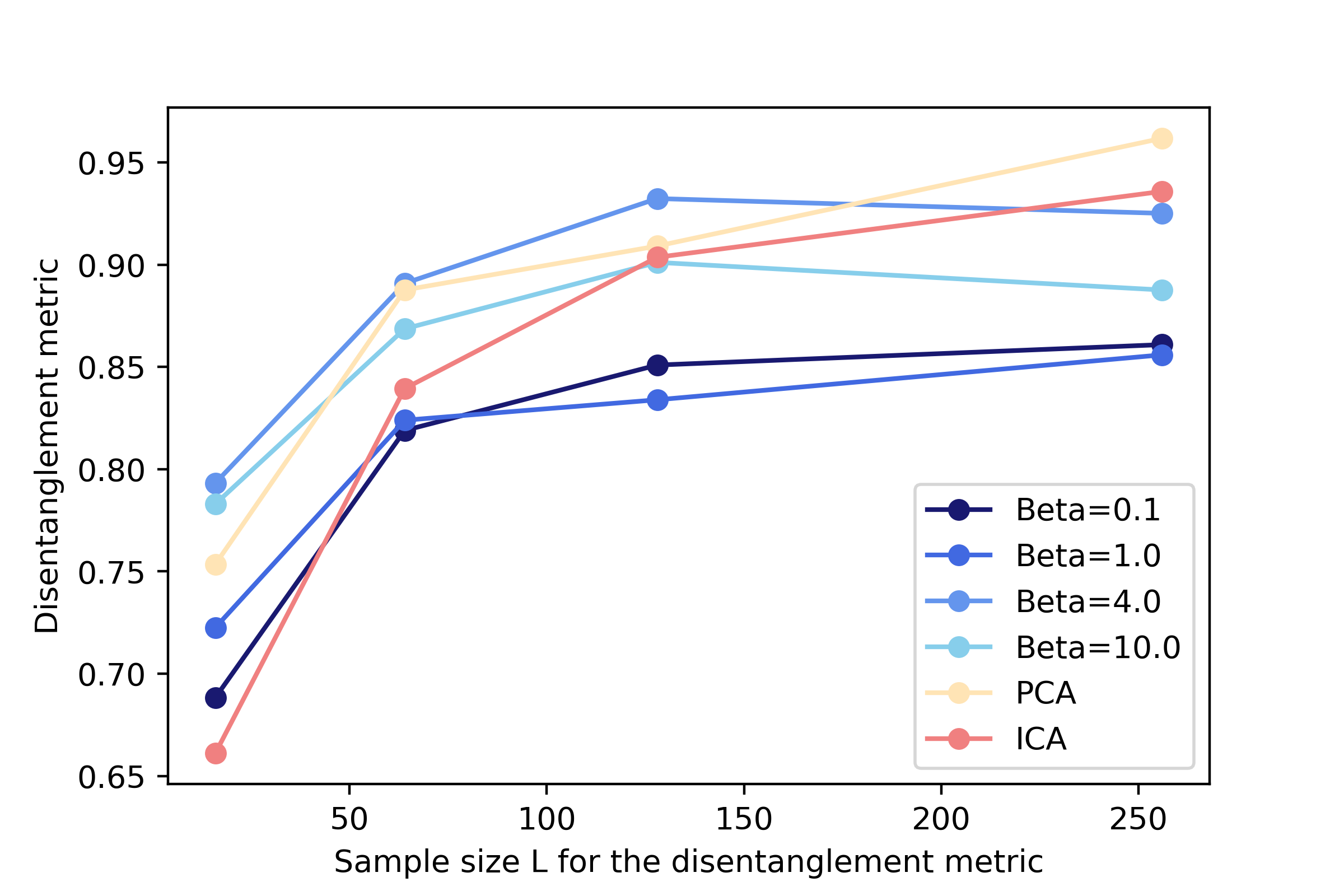}
\caption{Final disentanglement scores for various parameters of the disentanglement metric evaluated with the MLP VAE on 2Dshapes.}

\label{fig:sample_size}
\end{figure}

While the metric introduced by \cite{higgins2016beta} has a number of parameters, no guidance is provided on what values to choose. Follow-up work \cite{pmlr-v80-kim18b} notices a particular importance of the $L$ parameter which represents a sample size for generating the training samples. In Figure \ref{fig:sample_size}, we train the linear classifier for $L \in \{16,64,128,256\}$ and find that higher values improve the average disentanglement score significantly for PCA and ICA in particular. Moreover, it causes the $\beta$-VAEs to have increasingly similar performance which makes it harder to use the metric for model selection.

As seen in Table \ref{tab:linearvsnonlinear}, we also note that the choice of classifier does not have a significant effect on the disentanglement metric score. This contradicts the assertion of \cite{higgins2016beta}, who claim that using a non-linear classifier itself may disentangle the results.

\begin{table}[h]
    \centering
    
    \begin{tabularx}{0.981\textwidth}{l
                                    r
                                    r 
                                    r
                                    r 
                                    r
                                 }

        \toprule
         & \bf{Linear (PyTorch)}  &  \bf{MLP}   & \bf{Random Forest} &  \bf{Linear (SKlearn)}  \\
           \midrule
          2Dshapes & 77.62 $\pm$ 13.04\% & 78.8 $\pm$ 12.73\% & 80.16 $\pm$ 12.29\% & 80.73 $\pm$ 12.03\% \\
                                
          3Dshapes & 91.98 $\pm$ 6.82\% & 92.61 $\pm$ 6.82\% & 94.45 $\pm$ 6.52\% & 94.89 $\pm$ 5.67\% \\
          
          MPI3DToy & 33.21 $\pm$ 14.38\% & 34.78 $\pm$ 15.72\% & 36.25 $\pm$ 16.91\% & 35.86 $\pm$ 16.82\% \\
                                
            \bottomrule
    \end{tabularx}
    \caption{Disentanglement scores using different classifiers aggregated across $\beta$ values on 2Dshapes}
    \label{tab:linearvsnonlinear}
\end{table}

\subsection{Reconstruction}
\label{subsec:recon}



\begin{table}[h]
    \centering
    
    \begin{tabularx}{0.73\textwidth}{l
                                    r
                                    r 
                                    r
                                    r 
                                    r
                                 }

        \toprule
         & \bf{$0.1$-VAE}  &  \bf{$1$-VAE}   & \bf{$4$-VAE} &  \bf{$10$-VAE} &  \bf{WGAN-GP}  \\
           \midrule
          CIFAR10 & 197.8 & 171.1 & 221.1 & 266.5 & 29.3 \\
                                
          CIFAR100 & 132.1 & 163.5 & 204.6 & 252.9 & N/A\\
                                
            \bottomrule
    \end{tabularx}
    \caption{FID (lower is better) for different $\beta$ values on CIFAR10 and CIFAR100 compared with WGAN-GP \citep{NIPS2017_892c3b1c}.}
    \label{tab:CIFARFID}
\end{table}


Table ~\ref{tab:CIFARFID} shows the computed FID values for various values of the parameter $\beta$ and Figures \ref{fig:beta01recon}, \ref{fig:beta10recon} show the corresponding reconstructions. The already very high values of FID achieved by the standard VAE with $\beta=1$ steeply increase as we make the $\beta$ larger. Remarkably, CIFAR10 achieves the best FID for $\beta=1$, but CIFAR100 does so for $\beta=0.1$. \cite{rybkin2020simple} observe similar behaviour across datasets where small but nonzero $\beta$ tends to be the best, suggesting that $\beta$ cannot be seen as just a regularization hyperparameter given that it can improve the model performance even on training data. Interestingly, CIFAR100 consistently achieves lower FID scores than CIFAR10, even though CIFAR100 is a more complex dataset.

We have already shown in Figure \ref{fig:DM_Burgess} that for MPI3DToy, which is a dataset containing ground truth variation factors with complexity the closest to CIFAR10, the best values of disentanglement were also produced for $\beta=0.1$. We found the reconstructions for MPI3DToy to also be significantly lower quality than for 2Dshapes or 3Dshapes. This suggests that the model first needs to be able to produce good reconstructions, and then it can specialise in producing disentangled representations. There is likely to be a range of $\beta$ values for which both the reconstruction and disentanglement improve simultaneously, and only later there starts to be a trade-off between those two notions of representation quality. Based on this, we hypothesise that in order to scale disentanglement beyond toy datasets such as 2Dshapes, it is necessary to use models which are simultaneously very strong at image generation.

\begin{figure}[h]
  \centering
  \begin{minipage}[b]{0.49\textwidth}
    \includegraphics[width=\textwidth]{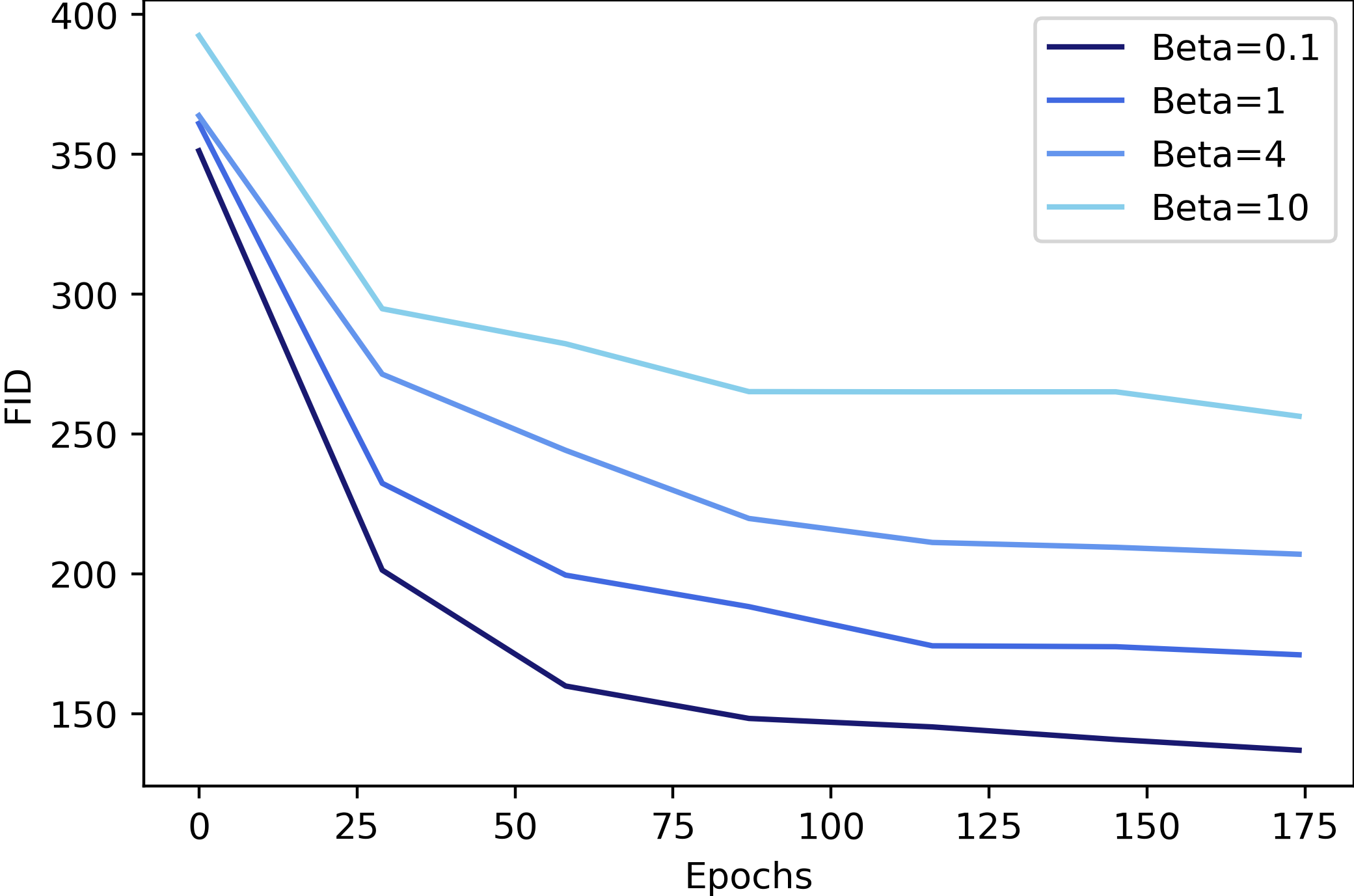}
    \caption{FID during training of CIFAR100 (for varying $\beta$)}
    \label{fig:MNISTFID}
  \end{minipage}
  \hfill
  \begin{minipage}[b]{0.49\textwidth}
    \includegraphics[width=\textwidth]{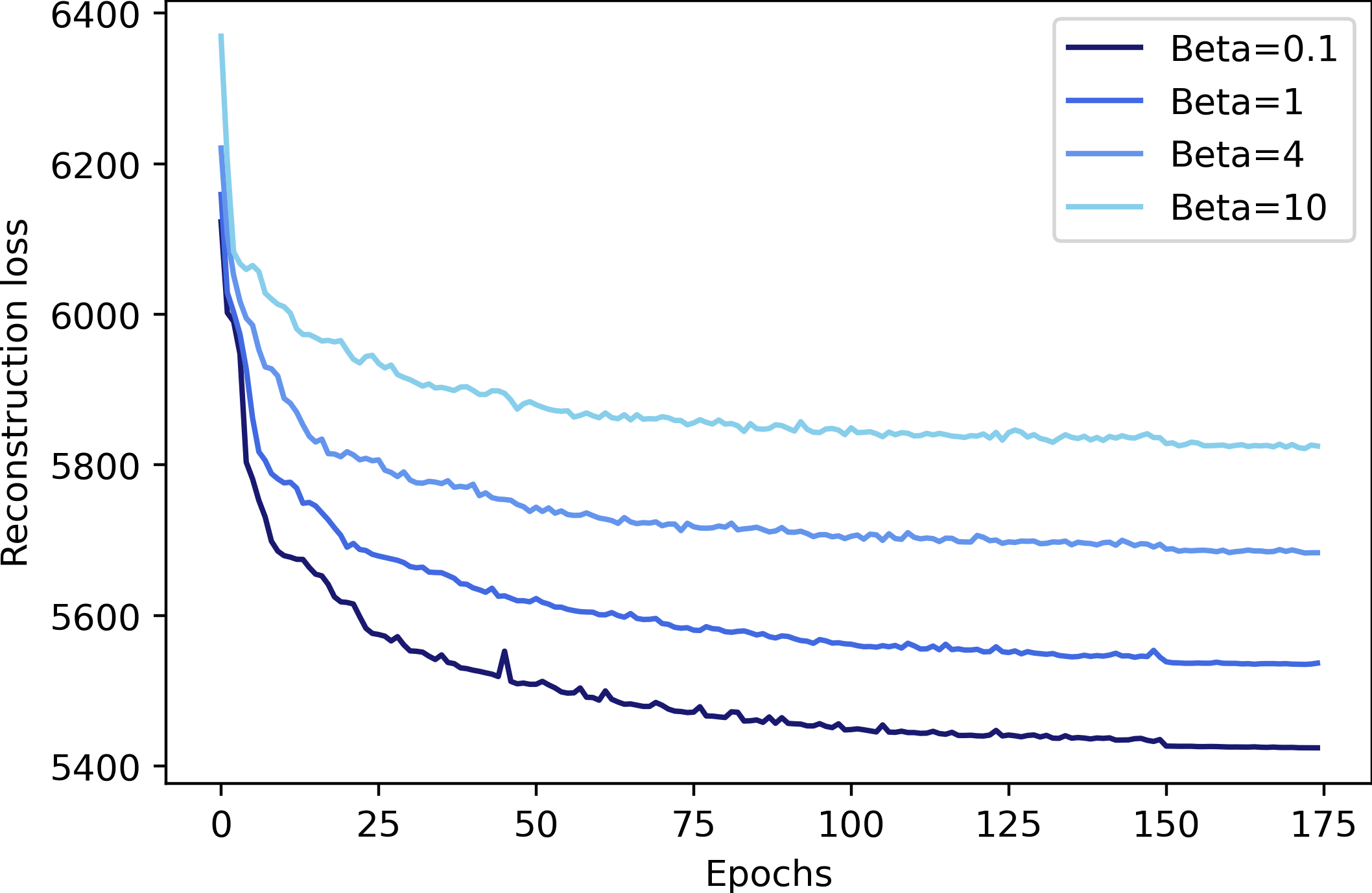}
    \caption{Reconstruction loss during CIFAR100 training (for varying $\beta$)}
    \label{fig:MNISTLoss}
  \end{minipage}
\end{figure}

\begin{figure}[!tbp]
  \centering
  \begin{minipage}[b]{0.47\textwidth}
    \includegraphics[width=\textwidth]{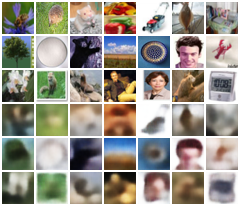}
    \caption{Reconstructions for CIFAR100 with $\beta=0.1$}
    \label{fig:beta01recon}
  \end{minipage}
  \hfill
  \begin{minipage}[b]{0.47\textwidth}
    \includegraphics[width=\textwidth]{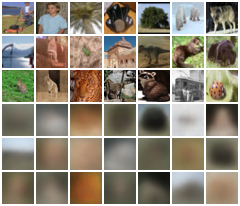}
    \caption{Reconstructions for CIFAR100 with $\beta=10$}
    \label{fig:beta10recon}
  \end{minipage}
\end{figure}

In Figure~\ref{fig:MNISTFID} and Figure~\ref{fig:MNISTLoss}, we show FID and reconstruction loss for training on CIFAR100. We found that even though VAEs tend to be evaluated by the negative log-likelihood, there is almost no qualitative difference between the behaviour of FID and NLL in our models \citep{dai2019diagnosing, lucic2017gans}. The reconstructions for the CIFAR datasets can be seen in Figure~\ref{fig:beta01recon} and Figure~\ref{fig:beta10recon}. It is clear that lower values of $\beta$ enable reconstructions that have a far higher quality, albeit ones that are still inferior to ones produced by state-of-the-art GANs, as measured by FID \citep{heusel2017gans, NIPS2017_892c3b1c}. Finally, we note that these qualitative inspections coincide with the quantitative evaluation, unlike the disentanglement metric.

\section{Discussion}
\subsection{Broad reproducibility}
Neither us nor follow-up work have been able to reproduce the original results of \cite{higgins2016beta}. The work by \cite{pmlr-v80-kim18b, locatello2020sober} was unable to significantly exceed 80$\%$ accuracy on 2Dshapes, regardless of the parameter setting. This contrasts with the 99$\%$ accuracy in the original paper. \cite{locatello2020sober} extensively show that even other metrics proposed in the literature exhibit high variance and inconsistency across datasets for the same model. In our results, it was crucial to both ignore the `shape' generative factor in 2Dshapes and to only consider median performance. Those things were only mentioned in the Appendix of the original paper but are vital for achieving close to the same performance.

\subsection{VAE with MSE loss}
\label{subsec:vae_mse}
$\beta$-VAE has a particular interpretation when assuming Gaussian likelihood over the decoded pixels. Using MSE loss is effectively equivalent to the normal VAE formulation but with a calibrated Normal prior. In the Gaussian likelihood case, we normally have $p(\vecb x|\vecb z) \sim N(\hat{\vecb x}, \vecb I)$ and the log-likelihood of the decoder is then 
\begin{equation}
    -\log p(\vecb x|\vecb z) = \frac{1}{2} \lVert\hat{\vecb x}-\vecb x\rVert^2 + D\log (\sqrt{2\pi}) = \frac{D}{2}\text{MSE}(\hat{\vecb x}, \vecb x) + c,
\end{equation}
where $\hat{\vecb x}$ is the prediction of the decoder and D is the dimensionality of the dataset. \cite{rybkin2020simple} show that if we instead assume $p(\vecb x|\vecb z) \sim N(\hat{\vecb x}, \sigma^2 \vecb I)$, the log likelihood would be 
\begin{equation}
    -\log p(\vecb x|\vecb z) = \frac{1}{2\sigma^2} \lVert\hat{\vecb x}-\vecb x\rVert^2 + D \log(\sigma) + c = \frac{D}{2\sigma^2}\text{MSE}(\hat{\vecb x}, \vecb x) + D \log\sigma + c
\end{equation}
and the full VAE objective is
\begin{equation}
    L = D \log(\sigma) + \frac{D}{2 \sigma^2} \text{MSE}(\hat{\vecb x}, \vecb x) + D_{KL}(q(\vecb z|\vecb x) \| p(\vecb z)).
\end{equation}
This is now very similar to the $\beta$-VAE objective since $\sigma^2$ plays the same weighting role as $\beta$, and if $\sigma^2$ is assumed to be constant, the $D\log(\sigma)$ term disappears during optimisation. This shows the equivalence of $\beta$-VAE to the standard VAE in this case.

Some of the experiments by \cite{higgins2016beta} use MSE loss and show results superior to the standard VAE. However, such examples cannot be used to show superiority of $\beta$-VAE as a model class since it becomes equivalent to the standard VAE with a different prior. 

\subsection{Implementation details}
We have initially adopted an online implementation of various VAE models \citep{yanndubs2019}, to which we added the disentanglement metric, more datasets, FID score, and more model options. In order to reproduce the results better, we later tried re-implementing the core model and training code while crosschecking it via other sources. The implementation of the disentanglement metric was initially constructed from scratch; we later tried another freely available implementation of the disentanglement metric \citep{noauthor_google-researchdisentanglement_lib_2021}, but achieved the same results. To ensure our implementation of FID was accurate, we based our code off the standard implementation by \cite{Seitzer2020FID}.

Our final code used to produce the results in this report is mostly custom-made and is available at \url{https://github.com/Mandelbrot99/BetaVAE}.

\section{Conclusion}
In this report, we studied the performance of $\beta$-VAE in terms of disentanglement and reconstruction across a variety of datasets. First, we observed that the results originally reported by \cite{higgins2016beta} are difficult to reproduce. They rely heavily on discarding of the worst 50$\%$ performing random seeds as well as simplifying the learning task, despite both of those aspects only being mentioned in the Appendix of the paper. The newly proposed disentanglement metric fails to fully capture human-interpretable disentanglement, as evidenced by qualitative evaluation. We show that even the hyperparameters of the disentanglement metric itself can be used to artificially boost the scores. 

On more complex datasets, we noted that the intuition suggested by \cite{higgins2016beta} of $\beta>1$ yielding better disentanglement scores breaks down. Rather, $\beta<1$ gives the best results on MPI3DToy. We hypothesise that this can be attributed to different relative magnitudes of the reconstruction loss and the KL divergence on more complex datasets. Finally, we confirmed that FID is strongly correlated with the reconstruction loss and that both notions coincide with a visual inspection of the reconstructed images. However, the best FID scores are not necessarily achieved for the lowest $\beta$ value which is somewhat counter-intuitive, similar to how the best disentanglement can be achieved for $\beta < 1$.

Our work could be further improved by running the experiments for many more random seeds than we did, but we were constrained by limited hardware. It is known that the variance of performance across random seeds and hyperparameters is generally even greater than across model choices \citep{locatello2020sober}. This weakens the robustness of our claims, given that we were not able to run very large-scale experiments. We would also be very interested in seeing more results on the disentanglement-reconstruction trade-off described in Section \ref{subsec:recon}, which would again require very extensive experiments with modern VAE-based models. For example, we would like to use the real MPI3D instead of the toy version to see if our results generalise to a truly real-world setting.

\bibliography{main}
\bibliographystyle{iclr2021_conference}

\renewcommand{\thesubsection}{\Alph{subsection}}
\appendix
\section*{Appendix}
\addcontentsline{toc}{section}{Appendix}

\subsection{Model and Hyperparameters Details}
\label{subsec:arch_details}
A summary of the model architectures we used is shown in Table \ref{tab:modelarchi}. The MLP model for 2Dshapes follows \cite{higgins2016beta} but the Convolutional VAE is adopted from \cite{burgess2018understanding} because other follow-up work typically uses the same architecture.

The settings of the optimiser likewise generally use the same parameters across all experiments except for 3Dshapes, where we found it necessary to decrease the learning rate to 1e-4. The 2Dshapes experiments use 256 batch size while the remaining datasets use 64. We also reduced the learning rate by a factor of 5 for last 25$\%$ epochs of training on all experiments.

All 2Dshapes experiments were ran using four seeds 123, 427, 235, 921. MPI3DToy and 3Dshapes used 123, 427, 235. CIFAR10 and CIFAR100 experiments used 4723, 1263.

\subsection{PCA and ICA implementation}
Because all common implementations of ICA and PCA are not able to train from minibatches and instead use expensive linear algebra routines with the full dataset, we had to significantly limit the training data size in order to execute the training. For PCA, we use 25000 samples for BW datasets and 3500 for RGB datasets while for ICA, we use 2500 for BW and 1000 for RGB datasets, respectively. Additionally, RGB images are flattened since standard PCA/ICA implementations are not able to support images with multiple channels.

\begin{table}[]
    \renewcommand*{\arraystretch}{2} 
    \centering
    \begin{tabular}{p{19mm} p{16mm} p{17mm} p{80mm}} 
         \Xhline{2\arrayrulewidth}
         \textbf{Dataset} & \textbf{Optimiser} & \textbf{} & \textbf{Architecture}  \\
         \Xhline{2\arrayrulewidth}
         
         2Dshapes (MLP) &
         Adagrad \newline 1e-2 \newline 256 &
         Input \newline Encoder \newline Latents \newline Decoder &
         4096 (flattened 64x64x1). \newline FC 1200, 1200. ReLU activation. \newline 10 \newline FC 1200, 1200, 1200, 4096. Tanh activation. Bernoulli. \\
         \hline
         
         2Dshapes (Conv VAE) &
         Adam \newline 5e-4 \newline 256 &
         Input \newline Encoder \newline Latents \newline Decoder &
         4096 (flattened 64x64x1). \newline Conv 32x4x4 (stride 2) 3x, FC 256 2x. ReLU activation. \newline 10 \newline Deconv reverse of encoder. ReLU activation. Bernoulli. \\
         \hline
         
         3Dshapes (Conv VAE) &
         Adam \newline 1e-4 \newline 64 &
         Input \newline Encoder \newline Latents \newline Decoder &
         12288 (flattened 64x64x3). \newline Conv 32x4x4 (stride 2) 3x, FC 256 2x. ReLU activation. \newline 10 \newline Deconv reverse of encoder. ReLU activation. Bernoulli. \\
         \hline
         
         MPI3dToy (Conv VAE) &
         Adam \newline 5e-4 \newline 64 &
         Input \newline Encoder \newline Latents \newline Decoder &
         12288 (flattened 64x64x3). \newline Conv 32x4x4 (stride 2) 3x, FC 256 2x. ReLU activation. \newline 10 \newline Deconv reverse of encoder. ReLU activation. Bernoulli. \\
         \hline
         
         CIFAR10 (Conv VAE) &
         Adam \newline 5e-4 \newline 64 &
         Input \newline Encoder \newline Latents \newline Decoder &
         12288 (flattened 64x64x3). \newline Conv 32x4x4 (stride 2) 3x, FC 256 2x. ReLU activation. \newline 128 \newline Deconv reverse of encoder. ReLU activation. Bernoulli. \\
         \hline
         
         CIFAR100 (Conv VAE) &
         Adam \newline 5e-4 \newline 64 &
         Input \newline Encoder \newline Latents \newline Decoder &
         12288 (flattened 64x64x3). \newline Conv 32x4x4 (stride 2) 3x, FC 256 2x. ReLU activation. \newline 128 \newline Deconv reverse of encoder. ReLU activation. Bernoulli. \\
         \hline
         
    \end{tabular}
    \caption{Model Architecture Details}
    \label{tab:modelarchi}
\end{table}

\end{document}